\pdfoutput=1

\documentclass[11pt]{article}

\usepackage[]{naacl2021}

\usepackage{times}
\usepackage{latexsym}

\usepackage[T1]{fontenc}

\usepackage[utf8]{inputenc}

\usepackage{microtype}

\usepackage{times}  
\usepackage{helvet} 
\usepackage{courier}  
\usepackage{graphicx} 
\usepackage{amsmath}
\usepackage{multirow}
\usepackage{lettrine}
\usepackage[switch]{lineno}
\usepackage{natbib}  
\usepackage{caption} 
\usepackage{color}
\definecolor{orange}{RGB}{237,124,55}
\definecolor{green}{RGB}{0,178,87}
\usepackage{url}

\title{ShadowGNN: Graph Projection Neural Network for Text-to-SQL Parser}

\author{Zhi Chen$^{1}$, Lu Chen$^{1}$\thanks{The corresponding authors are Lu Chen and Kai Yu.}, Yanbin Zhao$^{1}$, Ruisheng Cao$^{1}$, Zihan Xu$^{1}$,
\\ \textbf{Su Zhu}$^{2}$ \and \textbf{Kai Yu}$^{1}$\footnotemark[1] \\
        $^1$X-LANCE Lab, Department of Computer Science and Engineering \\ 
        MoE Key Lab of Artificial Intelligence, AI Institute, Shanghai Jiao Tong University\\
        Shanghai Jiao Tong University, Shanghai, China\\
    State Key Lab of Media Convergence Production Technology and Systems, Beijing, China\\
    $^2$AISpeech Co., Ltd., Suzhou, China \\
    \texttt{\{zhenchi713, chenlusz, kai.yu\}@sjtu.edu.cn}
        }
        
\begin{document}
\maketitle
\begin{abstract}
Given a database schema, Text-to-SQL aims to translate a natural language question into the corresponding SQL query. Under the setup of cross-domain, traditional semantic parsing models struggle to adapt to unseen database schemas. To improve the model generalization capability for rare and unseen schemas, we propose a new architecture, ShadowGNN, which processes schemas at abstract and semantic levels. By ignoring names of semantic items in databases, abstract schemas are exploited in a well-designed graph projection neural network to obtain delexicalized representation of question and schema. Based on the domain-independent representations, a relation-aware transformer is utilized to further extract logical linking between question and schema. Finally, a SQL decoder with context-free grammar is applied. On the challenging Text-to-SQL benchmark Spider, empirical results show that ShadowGNN outperforms state-of-the-art models. When the annotated data is extremely limited (only 10\% training set), ShadowGNN gets over absolute 5\% performance gain, which shows its powerful generalization ability. Our implementation will be open-sourced at \url{https://github.com/WowCZ/shadowgnn}.
\end{abstract}

\section{Introduction}

Recently, Text-to-SQL has drawn a great deal of attention from the semantic parsing community~\cite{berant2013semantic,rsc23-cao-acl19,cao2020unsupervised}. The ability to query a database with natural language (NL) engages the majority of users, who are not familiar with SQL language, in visiting large databases. A number of neural approaches have been proposed to translate questions into executable SQL queries. On public Text-to-SQL
\begin{figure}[t]
\centering
\includegraphics[width=0.5\textwidth]{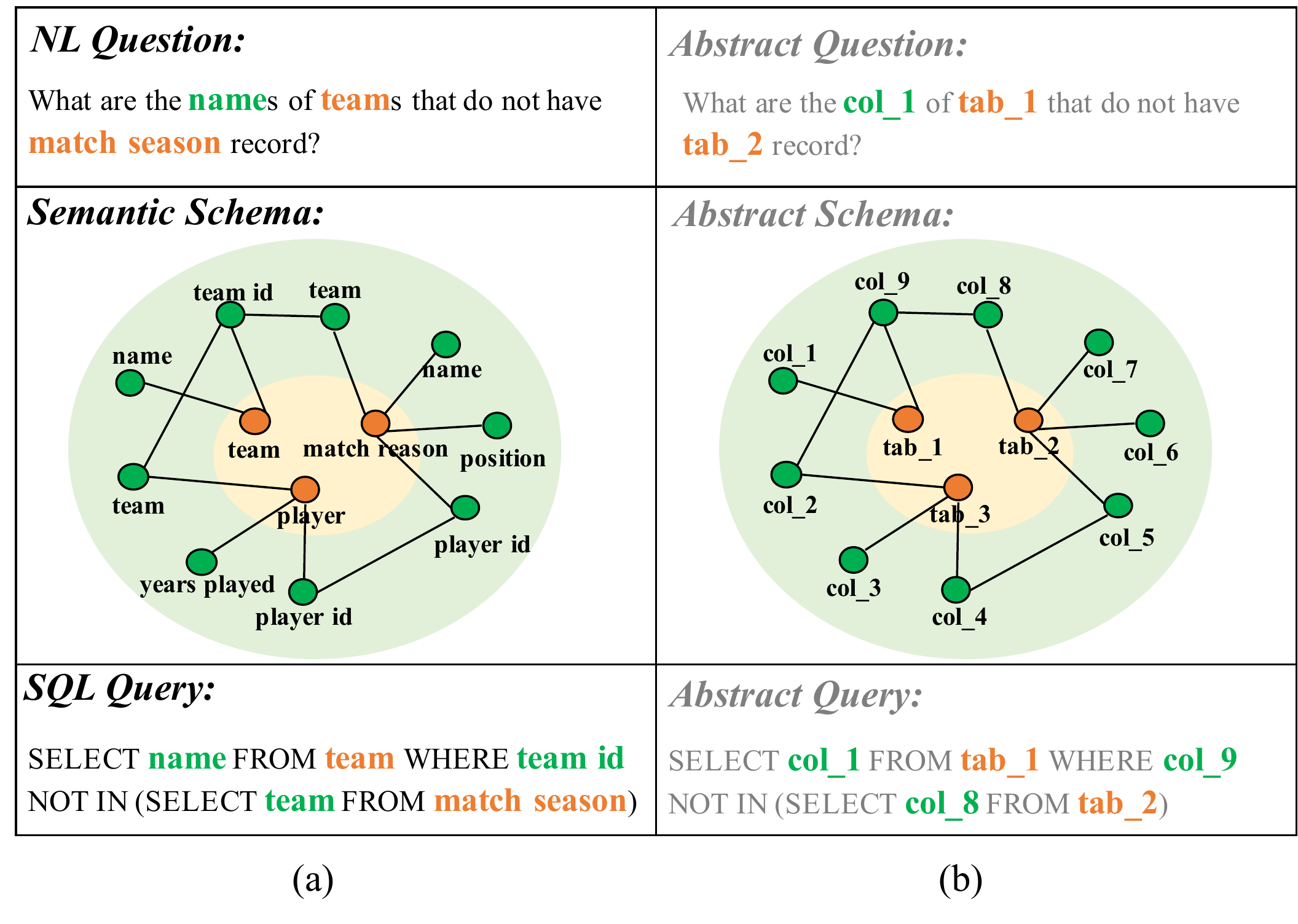} 
\vspace{-4mm}
\caption{An example to demonstrate the impact of domain information. (b) is the human-labeled abstract representation of Text-to-SQL content from domain-aware example (a). The \textcolor{green}{green} nodes and \textcolor{orange}{orange} nodes represent columns and tables respectively.}
\vspace{-4mm}
\label{fig:example}
\end{figure}
benchmarks~\cite{zhongSeq2SQL2017,krishnamurthy2017neural}, exact match accuracy even excesses more than 80\%. However, the cross-domain problem for Text-to-SQL is a practical challenge and ignored by the prior datasets. To be clarified, a database schema is regarded as a domain. 
The domain information consists of two parts: the semantic information (e.g., the table name) of the schema components and the structure information (e.g., the primary-key relation between a table and a column) of the schema.

The recently released dataset, Spider~\cite{yu2018spider}, hides the database schemas of the test set, which are totally unseen on the training set. In this cross-domain setup, domain adaptation is challenging for two main reasons. First, the semantic information of the domains in the test and development set are unseen in the training set. On the given development set, 35\% of words in database schemas do not occur in the schemas on the training set. It is hard to match the domain representations in the question and the schema. Second, there is a considerable discrepancy among the structure of the database schemas. Especially, the database schemas always contain semantic information. It is difficult to get the unified representation of the database schema. Under the cross-domain setup, the essential challenge is to alleviate the impact of the domain information.

First, it is necessary to figure out which role the semantic information of the schema components play during translating an NL question into a SQL query. Consider the example in Fig.~\ref{fig:example}(a), for the Text-to-SQL model, the basic task is to find out all the mentioned columns ($name$) and tables ($team$, $match$ $season$) by looking up the schema with semantic information (named as semantic schema). Once the mentioned columns and tables in the NL question are exactly matched with schema components, we can abstract the NL question and the semantic schema by replacing the general component type with the specific schema components. As shown in Fig.~\ref{fig:example}(b), we can still infer the structure of the SQL query using the abstract NL question and the schema structure. With the corresponding relation between semantic schema and abstract schema, we can restore the abstract query to executable SQL query with domain information. Inspired by this phenomenon, we decompose the encoder of the Text-to-SQL model into two modules. First, we propose a \textit{\textbf{G}raph \textbf{P}rojection \textbf{N}eural \textbf{N}etwork} (GPNN) to abstract the NL question and the semantic schema, where the domain information is removed as much as possible. Then, we use the relation-aware transformer to get unified representations of abstract NL question and abstract schema.

Our approach, named ShadowGNN, is evaluated on the challenging cross-domain Text-to-SQL dataset, Spider. Contributions are summarized as:
\begin{itemize}
\item We propose the ShadowGNN to alleviate the impact of the domain information by abstracting the representation of NL question and SQL query. It is a meaningful method to apply to similar cross-domain tasks.
\item To validate the generalization capability of our proposed ShadowGNN, we conduct the experiments with limited annotated data. The results show that our proposed  ShadowGNN can obtain absolute over 5\% accuracy gain compared with state-of-the-art model, when the annotated data only has the scale of 10\% of the training set.
\item The empirical results show that our approach outperforms state-of-the-art models (66.1\% accuracy on test set) on the challenging Spider benchmark. The ablation studies further confirm that GPNN is important to abstract the representation of the NL question and the schema.
\end{itemize}

\section{Background}
In this section, we first introduce relational graph convolution network (R-GCN)~\cite{schlichtkrull2018modeling}, which is the basis of our proposed GPNN. Then, we introduce the relation-aware transformer, which is a transformer variant considering relation information during calculating attention weights.
\subsection{Relational Graph Convolution Network}
Before describing the details of R-GCN, we first give notations of relational directed graph. We denote this kind of graph as $\mathcal{G} = (\mathcal{V}, \mathcal{E}, \mathcal{R})$ with nodes (schema components) $v_i \in \mathcal{V}$ and directed labeled edge $(v_i, r, v_j)\in \mathcal{E}$, where $v_i$ is the source node, $v_j$ is the destination node and $r \in \mathcal{R}$ is the edge type from $v_i$ to $v_j$. $\mathcal{N}_i^r$ represents the set of the neighbor indices of node $v_i$ under relation $r$, where $v_i$ plays the role of the destination node.

Each node of the graph has an input feature $\mathbf{x}_i$, which can be regarded as the initial hidden state $\mathbf{h}_i^{(0)}$ of the R-GCN. The hidden state of each node in the graph is updated layer by layer with following step:

\noindent \textbf{Sending Message} At the $l$-th layer R-GCN, each edge $(v_i, r, v_j)$ of the graph will send a message from the source node $v_i$ to the destination node $v_j$. The message is calculated as below:
\begin{equation}
\begin{aligned}
\mathbf{m}_{ij}^{(l)} = \mathbf{W}_r^{(l)}\mathbf{h}_i^{(l-1)},
\end{aligned}
\label{eq:send_message}
\end{equation}
where $r$ is the relation from $v_i$ to $v_j$ and $\mathbf{W}_r^{(l)}$ is a linear transformation, which is a trainable matrix. Following Equation~\ref{eq:send_message}, the scale of the parameter of calculating message is proportional to the number of the node types. To increase the scalability, R-GCN regularizes the message-calculating parameter with the basis decomposition method, which is defined as below:
\begin{equation}
\begin{aligned}
\mathbf{W}_r^{(l)} = \sum_{b=1}^B a_{rb}^{(l)}\mathbf{V}_b^{(l)},
\end{aligned}
\label{eq:basis_decomposition}
\end{equation}
where $B$ is the basis number, $a_{rb}^{(l)}$ is the coefficient of the basis transformation $\mathbf{V}_b^{(l)}$. For different edge types, the basis transformations are shared and only the coefficient $a_{rb}^{(l)}$ dependents on $r$.

\noindent \textbf{Aggregating Message} After the message sending process, all the incoming messages of each node will be aggregated. Combined with Equations~\ref{eq:send_message} and ~\ref{eq:basis_decomposition}, R-GCN simply averages these incoming messages as:
\begin{equation}
\begin{aligned}
\mathbf{g}_i^{(l)} &= \sum_{r \in \mathcal{R}} \sum_{j \in \mathcal{N}_i^r} \frac{1}{c_{i,r}} (\sum_{b=1}^B a_{rb}^{(l)}\mathbf{V}_b^{(l)}) \mathbf{h}_j^{(l-1)},
\end{aligned}
\label{eq:aggregation_mean}
\end{equation}
where $c_{i,r}$ equals to $|\mathcal{N}_i^r|$.

\noindent \textbf{Updating State} After aggregating messages, each node will update its hidden state from $\mathbf{h}_i^{(l-1)}$ to $\mathbf{h}_i^{(l)}$,
\begin{equation}
\begin{aligned}
\mathbf{h}_i^{(l)} = \sigma(\mathbf{g}_i^{(l)} + \mathbf{W}_0^{(l)}\mathbf{h}_i^{(l-1)}),
\end{aligned}
\label{eq:state_update}
\end{equation}
where $\sigma$ is an activation function (i.e., ReLU) and $\mathbf{W}_0^{(l)}$ is a weight matrix. For each layer of R-GCN, the update process can be simply denoted as:
\begin{equation}
\begin{aligned}
\mathbf{Y} = {\rm R}\textrm{-}{\rm GCN}(\mathbf{X}, \mathcal{G}),
\end{aligned}
\label{eq:rgcn}
\end{equation}
where $\mathbf{X} = \{\mathbf{h}_i\}_{i=1}^{|\mathcal{G}|}$, $|\mathcal{G}|$ is the number of the nodes and $\mathcal{G}$ is the graph structure.

\subsection{Relation-aware Transformer}
With the success of the large-scale language models, the transformer architecture has been widely used in natural language process (NLP) tasks to encode the sequence $X = [\mathbf{x}_i]_{i=1}^{n}$ with the \textit{self-attention} mechanism. As introduced in \citeauthor{vaswani2017attention}~\shortcite{vaswani2017attention},
a transformer is stacked by self-attention layers, where each layer transforms $\mathbf{x}_i$ to $\mathbf{y}_i$ with $H$ heads as follows:
\begin{align}
\label{eq:self_attention1}
&e_{ij}^{(h)} = \frac{\mathbf{x}_i\mathbf{W}_Q^{(h)}(\mathbf{x}_j\mathbf{W}_K^{(h)})^{\top}}{\sqrt{d_z/H}}, \\
&\alpha_{ij}^{(h)} = \mathop{\rm softmax}\limits_{j}\{e_{ij}^{(h)}\},  \\
\label{eq:self_attention2}
&\mathbf{z}_i^{(h)} = \sum_{j=1}^n \alpha_{ij}^{(h)} \mathbf{x}_j\mathbf{W}_V^{(h)},  \\
\label{eq:self_attention3}
&\mathbf{z}_i = \mathop{\rm Concat}(\mathbf{z}_i^{(1)}, \dots, \mathbf{z}_i^{(H)}), \\
&\bar{\mathbf{y}}_i = \mathop{\rm LayerNorm}(\mathbf{x}_i + \mathbf{z}_i),  \\
\label{eq:self_attention4}
&\mathbf{y}_i = \mathop{\rm LayerNorm}(\bar{\mathbf{y}}_i + \mathop{\rm FC}(\mathop{\rm ReLU}(\mathop{\rm FC}(\bar{\mathbf{y}}_i)))),
\end{align}
where $h$ is the head index, $d_z$ is the hidden dimension of $\mathbf{z}_i^{(h)}$, $\alpha_{ij}^{(h)}$ is attention probability, $\mathop{\rm Concat}$ denotes the concatenation operation, $\mathop{\rm LayerNorm}$ is layer normalization~\cite{ba2016layer} and $\mathop{\rm FC}$ is a full connected layer. The transformer function can be simply denoted as:
\begin{equation}
\begin{aligned}
\mathbf{Y} = \mathop{\rm Transformer}(\mathbf{X}),
\end{aligned}
\label{eq:transformer}
\end{equation}
where $\mathbf{Y} = \{\mathbf{y}_i\}_{i=1}^{|X|}$ and $\mathbf{X} = \{\mathbf{x}_i\}_{i=1}^{|X|}$ and $|X|$ is the sequence length.

Relation-aware transformer (RAT)~\cite{shaw2018self} is an important extension of the traditional transformer, which regards the input sequence as a labeled, directed, fully-connected graph. The pairwise relations between input elements are considered in RAT. RAT incorporates the relation information in Equation~\ref{eq:self_attention1}
and Equation~\ref{eq:self_attention2}. The edge from element $\mathbf{x}_i$ to element $\mathbf{x}_j$ is represented by vectors $\mathbf{r}_{ij,K}$ and $\mathbf{r}_{ij,V}$, which are represented as biases incorporated in self-attention layer, as follows:
\begin{align}
\label{eq:rat_self_attention1}
e_{ij}^{(h)} &= \frac{\mathbf{x}_i\mathbf{W}_Q^{(h)}(\mathbf{x}_j\mathbf{W}_K^{(h)} + \mathbf{r}_{ij,K})^{\top}}{\sqrt{d_z/H}}, \\ 
\alpha_{ij}^{(h)} &= \mathop{\rm softmax}\limits_{j}\{e_{ij}^{(h)}\}, \\
\label{eq:rat_self_attention2}
\mathbf{z}_i^{(h)} &= \sum_{j=1}^n \alpha_{ij}^{(h)} (\mathbf{x}_j\mathbf{W}_V^{(h)} + \mathbf{r}_{ij,V}),
\end{align}
where $\mathbf{r}_{ij,K}$ and $\mathbf{r}_{ij,V}$ are shared in different attention heads. For each layer of RAT, the update process can be simply represented as:
\begin{equation}
\begin{aligned}
\mathbf{Y} = \mathop{\rm RAT}(\mathbf{X}, \mathcal{R}),
\end{aligned}
\label{eq:rat}
\end{equation}
where $\mathcal{R} = \{R\}_{i=1, j=1}^{|X|, |X|}$ is the relation matrix among the sequence tokens and $R_{ij}$ means the relation type between $i$-th token and $j$-th token.

Both R-GCN and RAT have been successfully applied into Text-to-SQL tasks. \citet{bogin2019representing} utilizes R-GCN to encode the structure of the semantic schema to get the global representations of the nodes. \citet{rat-sql} considers not only the schema structure but also the schema link between the schema and the NL question. They proposed a unified framework to model the representation of the schema and the question with RAT. However, they do not explicitly explore the impact of the domain information. In the next section, we will introduce our proposed GPNN and explain how to use GPNN to get the abstract representation of the schema and the question.

\begin{figure*}[t]
\centering
\includegraphics[width=0.95\textwidth]{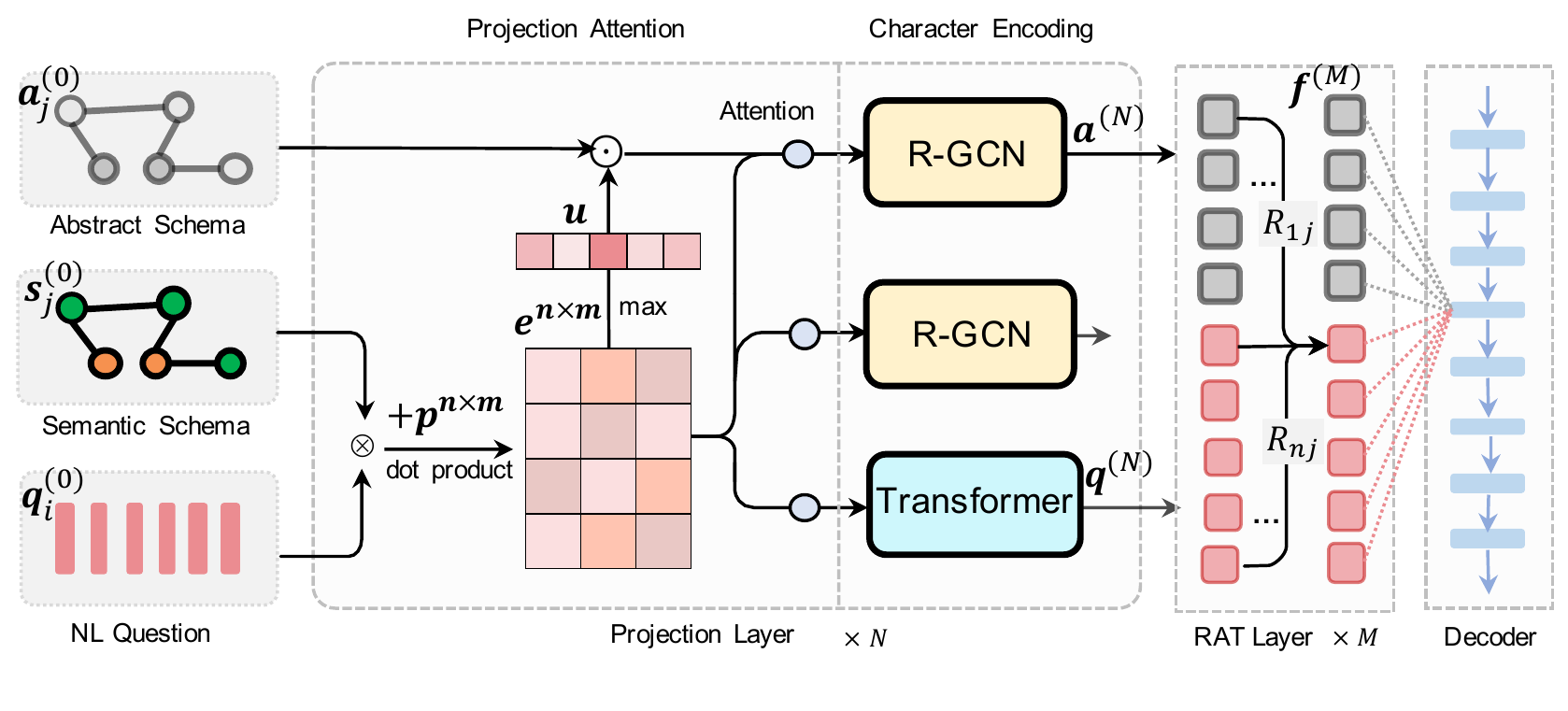} 
\vspace{-3mm}
\caption{The structure of our proposed ShadowGNN. ShadowGNN has three kinds of input: abstract schema, semantic schema, and natural language question. The encoder of ShadowGNN consists of two module: a stack of graph projection layers and a stack of relation-aware self-attention layers. To clarify the introduction of GPNN layer, we ignore the pretrained model RoBERTa in the figure.}
\vspace{-3mm}
\label{fig:model}
\end{figure*}

\section{Method}
\label{sec:method}
Text-to-SQL models take the NL questions $Q=\{q_i\}_{i=1}^n$ and the semantic schema $G=\{s_j\}_{j=1}^m$ as the input. In our proposed ShadowGNN, the encoder has been decomposed into two modules. The first module filters the specific domain information with a well-designed graph projection neural network (GPNN). The second module leverages relation-aware transformer to further get unified representations of question and schema. This two-phase encoder of ShadowGNN simulates the inference process of a human when translating a question to a SQL query under cross-domain setup: abstracting and inferring.

\subsection{Graph Projection Neural Network}
In this subsection, we introduce the structure of GPNN. As we discussed, the schema consists of database structure information and domain semantic information. GPNN looks at the schema from these two perspectives. Thus, GPNN has three kinds of inputs, abstract schema, semantic schema, and NL question. The input of the abstract schema is the type (table or column) of the schema nodes without any domain information, which can be regarded as a projection of semantic schema. Each node in the abstract schema is represented by a one-hot vector $\mathbf{a}_j^{(0)}$, which has two dimensions.

For semantic schema and NL question, we first use pretrained language model RoBERTa~\cite{liu2019roberta} to initialize their representations. We directly concatenate NL question and semantic schema together, which formats as `` [CLS] question [SEP] tables columns [SEP]". Each node name in the semantic schema may be tokenized into several sub-tokens or sub-words. We add an average pooling layer behind the final layer of the RoBERTa to align the sub-tokens to the corresponding node. We indicate the initial representation of NL question and semantic schema as $\mathbf{q}_i^{(0)}$ and $\mathbf{s}_j^{(0)}$. 

The main motivation of GPNN is to abstract the representations of question and schema. The abstract schema has been distilled from the semantic schema. The essential challenge lies on abstracting question representation. There are two separate operations in each GPNN layer: \textbf{Projection Attention} and \textbf{Character Encoding}. The projection attention of GPNN is to take the semantic schema as the bridge, where question updates its representation using abstract schema but attention information is calculated with the vectors of semantic schema. The character encoding is to augment the structure representation of the question sentence and the schema graph.

\noindent \textbf{Projection Attention} In each GPNN layer, there is first an attention operation between NL question and semantic schema, as follows:
\begin{align}
\label{eq:gpnn_attention1}
e_{ij} &= \mathbf{q}_i^{(l)}\mathbf{W}_Q^{(l)}(\mathbf{s}^{(l)}_j\mathbf{W}_K^{(l)})^{\top}, \\
\label{eq:gpnn_attention21}
\alpha_{ij} &= \mathop{\rm softmax}\limits_{j}\{e_{ij}\},
\end{align}
where $\mathbf{W}_Q^{(l)}$ and $\mathbf{W}_K^{(l)}$ are trainable parameters at $l$-th projection layer and $\mathbf{e}^{n \times m} = \{e_{ij}\}_{i=1,j=1}^{n,m}$ is the matrix of the weight score. $n$ is the length of the question, and $m$ is the number of schema nodes.

Before operating attention mechanism, inspired by \cite{bogin2019representing}, we first calculate the maximum values $\mathbf{u}$ of attention probability,
\begin{align}
\label{eq:max_score}
\mathbf{u}_j = \mathop{\rm max}\limits_{i}\{\alpha_{ij}\},
\end{align}
where the physical meaning of $\mathbf{u}_j$ is the most probability that the $j$-th component of the schema is mentioned by the question. We distinct the initial representation of the abstract schema by multiplying $\mathbf{u}$ on $l$-th layer abstract schema representation $\mathbf{a}^{(l)}$ in element-wise way, $\hat{\mathbf{a}}^{(l)} = \mathbf{a}^{(l)}\cdot\mathbf{u}$.

When updating the question representation, we take the representation of augmented abstract schema $\hat{\mathbf{a}}^{(l)}$ as key value of attention at $l$-th layer of GPNN,
\begin{align}
\label{eq:gpnn_attention22}
\mathbf{b}_i =& \sum_{j=1}^m \alpha_{ij} \hat{\mathbf{a}}^{(l)}_j\mathbf{W}_V^{(l)}, \\
\label{eq:gpnn_attention23}
\bar{\mathbf{q}}_i^{(l+1)} = \mathop{\rm gate}(\mathbf{b}_i) &* \mathbf{b}_i + (1-\mathop{\rm gate}(\mathbf{b}_i))*\mathbf{q}_i^{(l)},
\end{align}
where $\mathop{\rm gate}(\cdot) = \mathop{\rm sigmoid}(\mathop{\rm Linear}(\cdot))$ and $\mathbf{W}_V^{(l)}$ is trainable weight. When updating semantic schema, we take the transpose of the above attention matrix as the attention from schema to question,
\begin{align}
\label{eq:gpnn_attention3}
\mathbf{\hat{e}}^{m \times n} = (\mathbf{e}^{n \times m})^{\top} = \{\hat{e}_{ij}\}_{i=1,j=1}^{m,n}.
\end{align}
Similar to the update process of question from Equation~\ref{eq:gpnn_attention1}-~\ref{eq:gpnn_attention23}, the update process of semantic schema $\bar{\mathbf{s}}^{(l+1)}$ takes $\mathbf{\hat{e}}^{m \times n}$ as attention score and $\mathbf{q}^{(l)}$ as attention value. 
We can see that we only use the augmented abstract schema to update the question representation. In this way, the domain information contained in question representation will be removed. The update process of the abstract schema $\bar{\mathbf{a}}^{(l+1)}$ is the same as the semantic schema updating, where their attention weight $\mathbf{\hat{e}}^{m \times n}$ on the question $\mathbf{q}^{(l)}$ is shared. Noting that the input of attention operation for the abstract schema is the augmented abstract representation $\hat{\mathbf{a}}$.

\noindent \textbf{Character Encoding} We have used the projection attention mechanism to update the three kinds of vectors. Then, we combine the characters of schema and NL question and continue encoding schema and question with ${\rm R}\textrm{-}{\rm GCN}(\cdot)$ function and $\mathop{\rm Transformer}(\cdot)$ function respectively, as shown in Fig.~\ref{fig:model}.,
\begin{align}
\mathbf{a}^{(l+1)} &= \mathop{\rm R\textrm{-}GCN}(\bar{\mathbf{a}}^{(l+1)}, G), \\
\mathbf{s}^{(l+1)} &= \mathop{\rm R\textrm{-}GCN}(\bar{\mathbf{s}}^{(l+1)}, G), \\
\mathbf{q}^{(l+1)} &= \mathop{\rm Transformer}(\bar{\mathbf{q}}^{(l+1)}).
\end{align}
Until now, the projection layer has been introduced. Graph projection neural network (GPNN) is a stack of the projection layers. After GPNN module, we get the abstract representation of the schema and the question, indicated as $\mathbf{a}^{(N)}$ and $\mathbf{q}^{(N)}$.

\subsection{Schema Linking}
The schema linking~\cite{guo2019towards,lei-etal-2020-examining} can be regarded as a kind of prior knowledge, where the related representation between question and schema will be tagged according to the matching degree. There are 7 tags in total: Table Exact Match, Table Partial Match, Column Exact Match, Column Partial Match, Column Value Exact Match, Column Value Partial Match, and No Match. The column values store in the databases. As the above description, the schema linking can be represented as $\mathcal{D}=\{d_{ij}\}_{i=1,j=1}^{n, m}$, which $d_{ij}$ means the match degree between $i$-th word of question and $j$-th node name of schema. To integrate the schema linking information into GPNN module, we calculate a prior attention score $\mathbf{p}^{n \times m}={\rm Linear}({\rm Embedding}(\mathbf{d}_{ij}))$, where $\mathbf{d}_{ij}$ is the one-hot representation of match type $d_{ij}$. The attention score in Equation~\ref{eq:gpnn_attention1} is updated as following:
\begin{align}
\label{eq:gpnn_attention_new}
e_{ij} &= \mathbf{q}_i^{(l)}\mathbf{W}_Q^{(l)}(\mathbf{s}^{(l)}_j\mathbf{W}_K^{(l)})^{\top} + p_{ij},
\end{align}
where $p_{ij}$ is the prior score from $\mathbf{p}^{n \times m}$. The prior attention score is shared among all the GPNN layers.

\subsection{RAT}
If we split the schema into the tables and the columns, there are three kinds of inputs: question, table, column. RATSQL~\cite{rat-sql} leverages the relation-aware transformer to unify the representation of the three inputs. RATSQL defines all the relations $\mathcal{R}=\{R_{ij}\}_{i=1,j=1}^{(n+m), (n+m)}$ among the three inputs and uses the $\mathop{\rm RAT}(\cdot)$ function to get unified representation of question and schema. The details of the defined relations among three components are introduced in RATSQL~\cite{rat-sql}. The schema linking relations are the subset of $\mathcal{R}$. In this paper, we leverage the RAT to further unify the abstract representation of question $\mathbf{q}^{(N)}$ and schema $\mathbf{a}^{(N)}$, which is generated by previous GPNN module. We concatenate sentence sequence $\mathbf{q}^{(N)}$ and schema sequence $\mathbf{a}^{(N)}$ together into a longer sequence representation, which is the initial input of RAT module. After RAT module, the final unified representation of question and schema is indicated as:
\begin{align}
\mathbf{f}^{(M)} &= \mathop{\rm RAT}({\rm concat}(\mathbf{q}^{(N)}, \mathbf{a}^{(N)}), \mathcal{R}).
\end{align}

\subsection{Decoder with SemQL Grammar}
To effectively constrain the search space during synthesis, IRNet~\cite{guo2019towards} designed a context-free SemQL grammar as the intermediate representation between NL question and SQL, which is essentially an abstract syntax tree (AST). SemQL recovers the tree nature of SQL. To simplify the grammar tree, SemQL in IRNet did not cover all the keywords of SQL. For example, the columns contained in $\mathop{\rm GROUPBY}$ clause can be inferred from $\mathop{\rm SELECT}$ clause or the primary key of a table where an aggregate function is applied to one of its columns. In our system, we improve the SemQL grammar, where each keyword in SQL sentence is corresponded to a SemQL node. During the training process, the labeled SQL needs to be transferred into an AST. During the evaluation process, the AST needs to recovered as the corresponding SQL. The recover success rate means the rate that the recovered SQL totally equals to labeled SQL. Our improved grammar raises the recover success rate from 89.6\% to 99.9\% tested on dev set.

We leverage the coarse-to-fine approach~\cite{dong2018coarse} to decompose the decoding process of a SemQL query into two stages, which is similar with IRNet. The first stage is to predict a skeleton of the SemQL query with skeleton decoder. Then, a detail decoder fills in the missing details in the skeleton by selecting columns and tables.

\section{Experiments}
In this section, we evaluate the effectiveness of our proposed ShadowGNN than other strong baselines. We further conduct the experiments with limited annotated training data to validate the generalization capability of the proposed ShadowGNN. Finally, we ablate other designed choices to understand their contributions.

\subsection{Experiment Setup}
\noindent \textbf{Dataset \& Metrics} We conduct the experiments on the Spider~\cite{yu2018spider}, which is a large-scale, complex and cross-domain Text-to-SQL benchmark. The databases on the Spider are split into 146 training, 20 development and 40 test. The human-labeled question-SQL query pairs are divided into 8625/1034/2147 for train/development/test. The test set is not available for the public, like all the competition challenges. We report the results with the same metrics as~\cite{yu2018spider}: exact match accuracy and component match accuracy.

\noindent \textbf{Baselines} The main contribution of this paper lies on the encoder of the Text-to-SQL model. As for the decoder of our evaluated models, we improve the SemQL grammar of the IRNet~\cite{guo2019towards}, where the recover success rate raises from 89.6\% to 99.9\%. The SQL query first is represented by an abstract syntax tree (AST) following the well-designed grammar~\cite{lin2019grammar}. Then, the AST is flattened as a sequence (named SemQL query) by the deep-first search (DFS) method. During decoding, it is still predicted one by one with LSTM decoder. We also leverage the coarse-to-fine approach to the decoder as IRNet. A skeleton decoder first outputs a skeleton of the SemQL query. Then, a detail decoder fills in the missing details in the skeleton by selecting columns and tables. R-GCN~\cite{bogin2019representing,kelkar2020bertrand} and RATSQL~\cite{rat-sql} are two other strong baselines, which improve the representation ability of the encoder.

\noindent \textbf{Implementations}
We implement ShadowGNN and our baseline approaches with PyTorch~\cite{paszke2019pytorch}. We use the pretrained models RoBERTa from PyTorch transformer repository~\cite{Wolf2019HuggingFacesTS}. We use Adam with default hyperparameters for optimization. The learning rate is set to 2e-4, but there is 0.1 weight decay for the learning rate of pretrained model. The hidden sizes of GPNN layer and RAT layer are set to 512. The dropout rate is 0.3. Batch size is set to 16. The layers of GPNN and RAT in ShadowGNN encoder are set to 4.

\begin{table}[t]
\centering
\resizebox{\columnwidth}{!}{
\begin{tabular}{c|c|c}
    \hline
    \textbf{Approaches}  & \textbf{Dev.} & \textbf{Test} \\
    \hline
    \hline
    Global-GNN~\cite{bogin2019global} & 52.7\% & 47.4\%  \\
    \hline
    R-GCN + Bertrand-DR~\cite{kelkar2020bertrand} & 57.9\% & 54.6\% \\
    \hline
    IRNet v2~\cite{guo2019towards} & 63.9\% & 55.0\% \\
    \hline
    RATSQL v3 + BERT-large~\cite{rat-sql} & 69.7\% & 65.6\% \\
    \hline
    \hline
    RATSQL$^\clubsuit$ + RoBERTa-large & 70.2\% & 64.0\% \\
    \hline
    GPNN + RoBERTa-large & 69.9\% & 65.7\% \\
    \hline
    ShadowGNN + RoBERTa-large & \textbf{72.3\%} & \textbf{66.1}\% \\
    \hline
\end{tabular}}
\caption{The exact match accuracy on the development set and test set. $^\clubsuit$ means the model is implemented by us, where the only difference is the encoder part compared with the proposed ShadowGNN model.}
\label{tab:table1}
\end{table}

\subsection{Experimental Results}
To fairly compared with our proposed ShadowGNN, we
implement RATSQL~\cite{rat-sql} with the same coarse-to-fine decoder and RoBERTa augmentation of ShadowGNN model. We also report the performance of GPNN encoder on test set. The detail implementations of these two baselines show as following:

\begin{itemize}
\item \noindent \textbf{RATSQL$^\clubsuit$} RATSQL model replaces the four projection layers with another four relation-aware self-attention layers. There are totally eight relation-aware self-attention layers in the encoder, which is consistent with orignal RAT-SQL setup~\cite{rat-sql}.
\item \noindent \textbf{GPNN} Compared with ShadowGNN, GPNN model directly removes the relation-aware transformer. There are only four projection layers in the encoder, which can get better performance than eight layers.

\end{itemize}
\begin{figure}[t]
\centering
\includegraphics[width=0.5\textwidth]{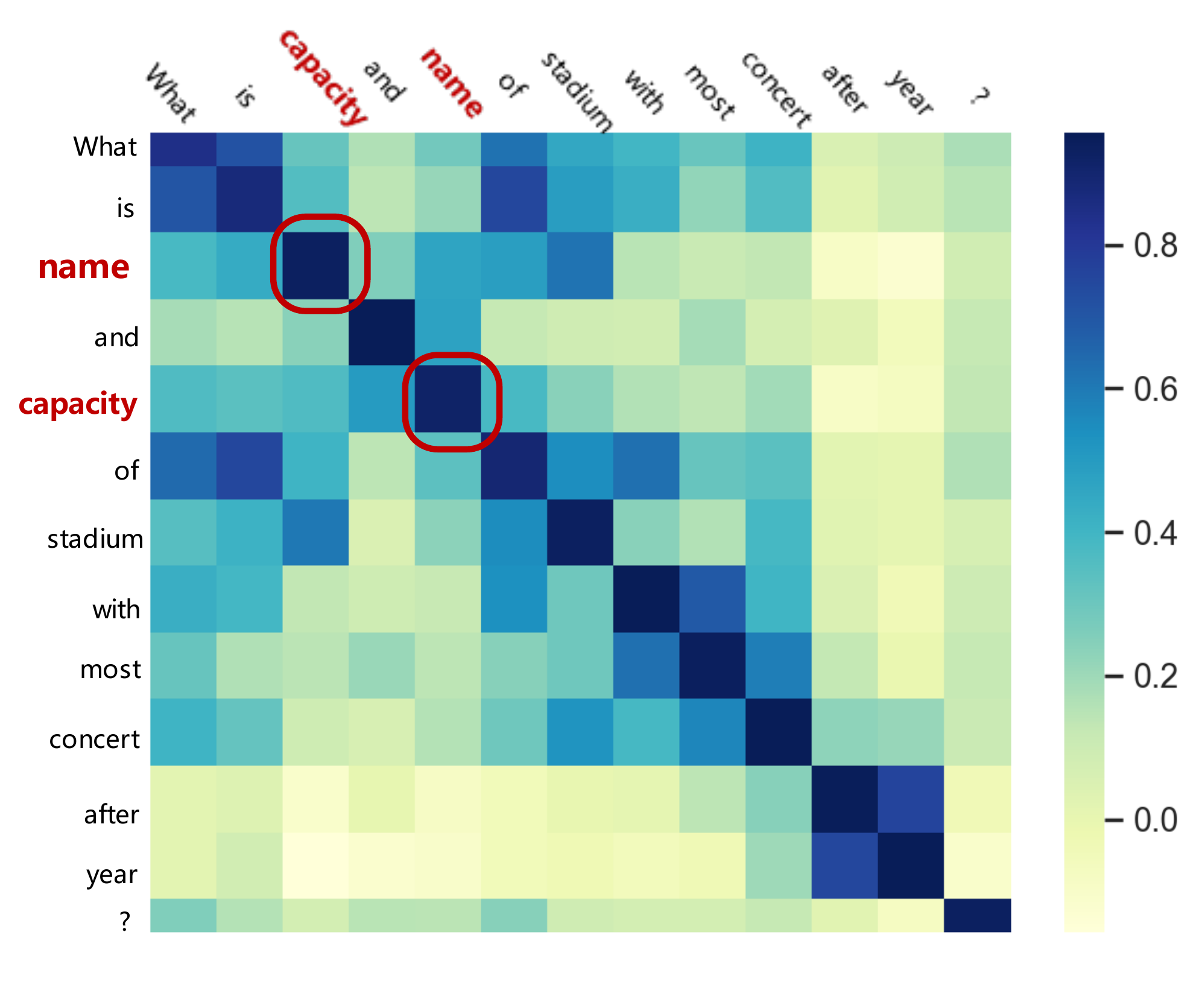} 
\caption{The cosine similarity of two questions. The positions of ``name" and 'capacity' in the two questions are exchanged.}
\label{fig:hotmap}
\end{figure}

Table~\ref{tab:table1} presents the exact match accuracy of the novel models on development set and test set. Compared with the state-of-the-art RATSQL, our proposed ShadowGNN gets absolute 2.6\% and 0.5\% improvement on development set and test set with RoBERTa augmentation. Compared with our implemented RATSQL$^\clubsuit$, ShadowGNN can still stay ahead, which has absolute 2.1\% and 2.1\% improvement on development set and test set. ShadowGNN improved the encoder and SemQL grammar of IRNet obtains absolute 11.1\% accuracy gain on test set. As shown in Table~\ref{tab:table1}, our proposed pure GPNN model achieves comparable performance with state-of-the-art approach on test set. Compared with other GNN-based models (Global-GNN and R-GCN), GPNN gets over 10\% improvement on development set and test set. To the best of our knowledge, our proposed GPNN gets the best performance on Spider dataset among all the GNN-based models. 

\subsection{Generalization Capability}
We design an experiment to validate the effectiveness of the graph projection neural network (GPNN). Considering a question ``What is name and capacity of stadium with most concert after year ?", which has been preprocessed, ``name" and ``capacity" are column names. We exchange their positions and calculate the cosine similarity with the representations of the final GPNN layer in ShadowGNN model. Interestingly, we find that ``name" has the most similar with ``capacity", as shown in Figure~\ref{fig:hotmap}. The semantic meaning of the two column names seems to be removed that the representations of the two column names only dependent on the existed positions. It indicates the GPNN can get the abstract representation of the question.

\begin{figure}[t]
\centering
\includegraphics[width=0.5\textwidth]{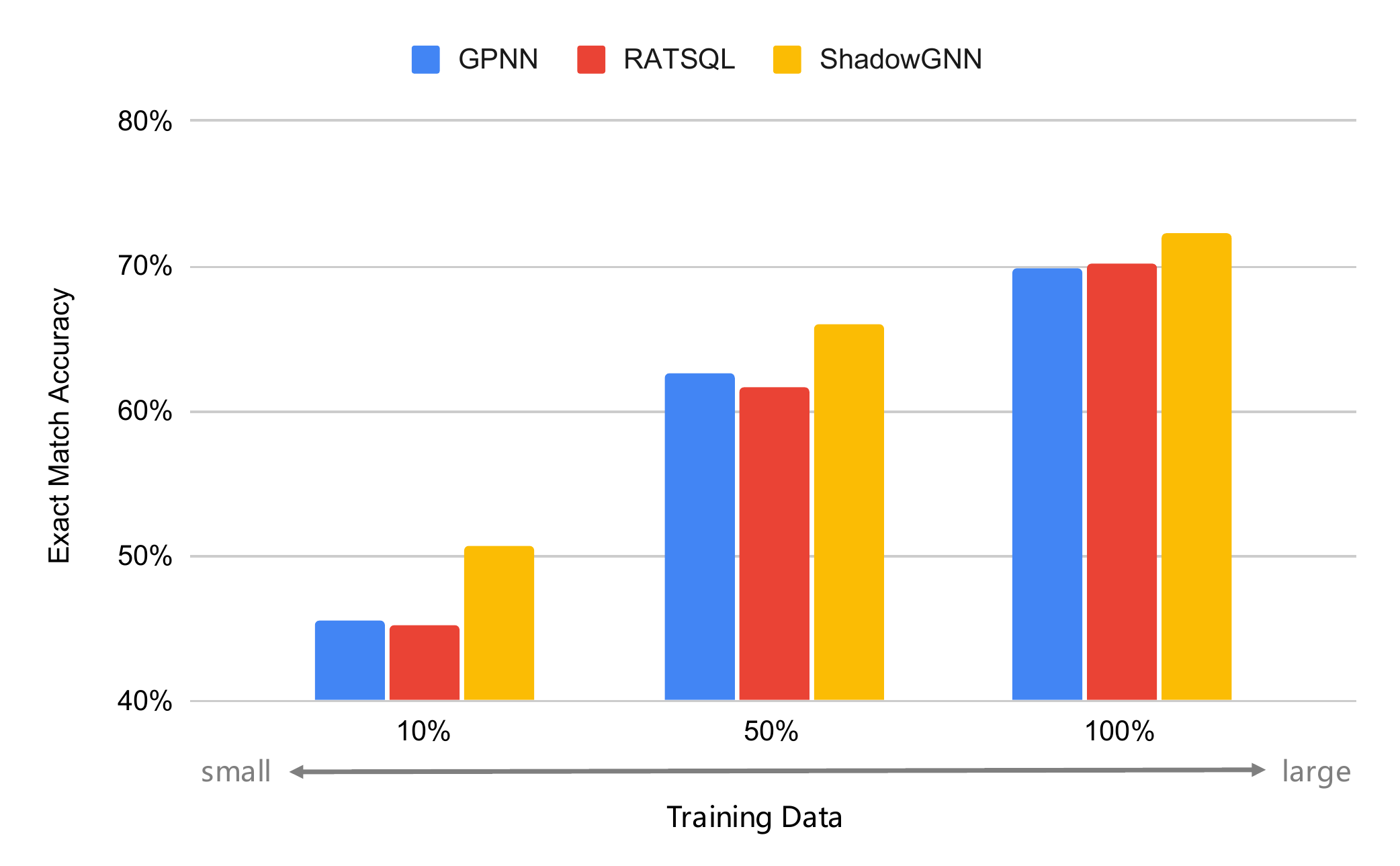} 
\caption{The exact match accuracy of GPNN, RATSQL and ShadowGNN on the limited training datasets. The limited training datasets are randomly sampled from fully training dataset with 10\%, 50\% and 100\% sampling probability.}
\label{fig:limit}
\end{figure}

To further validate the generalization ability of our proposed ShadowGNN, we conduct the experiments on the limited annotated training datasets. The limited training datasets are sampled from fully training dataset with 10\%, 50\% and 100\% sampling rate. As shown in Figure~\ref{fig:limit}, there is a large performance gap between RATSQL and ShadowGNN, when the annotated data is extremely limited only occupied 10\% of the fully training dataset. ShadowGNN outperforms RATSQL and GPNN with over 5\% accuracy rate on development set. Under this limited training data setup, we find an interesting phenomenon that the convergence speed of ShadowGNN is much faster than the other two models. As described in Section~\ref{sec:method}, the two-phase encoder of ShadowGNN simulates the inference process of a human when translating a question to a SQL query: abstracting and inferring. The experiments on limited annotated training datasets show these two phases are both necessary, which not only can improve the performance but also speed up the convergence.

\subsection{Ablation Studies}
We conduct ablation studies to analyze the contributions of well-designed graph projection neural network (GPNN). Except RATSQL and GPNN models, we implement other two ablation models: R-GCN and R-GCN+RAT. First, we introduce the implementations of the ablation models.

\begin{itemize}
\item \noindent \textbf{R-GCN$^\clubsuit$} We directly remove the projection part in the GPNN. When updating the question representation, we use the representation of semantic schema as attention value instead of abstract representation.
\item \noindent \textbf{R-GCN+RAT} In this model, there are four R-GCN layers and four relation-aware self-attention layers. To be comparable, the initial input of R-GCN is the sum of semantic schema and abstract schema.
\end{itemize}

\begin{table}[t]
\centering
\resizebox{\columnwidth}{!}{
\begin{tabular}{c|c|c|c|c|c}
    \hline
    \textbf{Approaches} & \textbf{Easy} & \textbf{Medium} & \textbf{Hard} & \textbf{Extra Hard} & \textbf{All} \\
    \hline
    \hline
     R-GCN~\cite{kelkar2020bertrand} & 70.4\% & 54.1\% & 35.6\% & 28.2\% & 50.7\%  \\
    \hline
     R-GCN$^\clubsuit$ & 78.9\% & 63.2\% & 46.6\% & 29.8\% & 58.7\%  \\
    \hline
     R-GCN+RAT & 85.0\% & 70.9\% & 56.3\% & 32.7\% & 65.6\%  \\
    \hline
     GPNN & \textbf{87.5}\% & 74.9\% & 59.2\% & 41.6\%  & 69.9\%  \\
    \hline
     RATSQL$^\clubsuit$ & 87.1\% & 74.9\% & 57.5\% & \textbf{46.4}\% & 70.2\%  \\
    \hline
     ShadowGNN & \textbf{87.5}\% & \textbf{78.0}\% & \textbf{61.5}\% & 45.8\% & \textbf{72.3}\%  \\
    \hline
\end{tabular}}
\caption{The match accuracy of the ablation methods at four hardness levels on development set. $^\clubsuit$ means the model is implemented by us.}
\label{tab:table2}
\end{table}

The decoder parts of these four ablation models are the same as the decoder of ShadowGNN. We present the accuracy of the ablation models at the four hardness levels on the development set, which is defined in \cite{yu2018spider}. As shown in Table~\ref{tab:table2}, ShadowGNN can get the best performance at three hardness levels. Compared with R-GCN~\cite{kelkar2020bertrand}, our implemented R-GCN based on SemQL grammar gets higher performance. Compared with R-GCN+RAT model, ShadowGNN still gets the better performance, where the initial input information is absolutely the same. It denotes that it is necessary and effective to abstract the representation of question and schema explicitly.


\section{Related Work}

\noindent \textbf{Text-to-SQL}
Recent models evaluated on Spider have pointed out several interesting directions for Text-to-SQL research. An AST-based decoder~\cite{yin2017syntactic} was first proposed for generating general-purpose programming languages. IRNet~\cite{guo2019towards} used a similar AST-based decoder to decode a more abstracted intermediate representation (IR), which is then transformed into an SQL query. RAT-SQL~\cite{rat-sql} introduced a relation-aware transformer encoder to improve the joint encoding of question and schema, and reached the best performance on the Spider~\cite{yu2018spider} dataset. BRIDGE~\cite{lin-etal-2020-bridging} leverages the database content to augment the schema representation. RYANSQL~\cite{choi2020ryansql} formulates the Text-to-SQL task as a slot-filling task to predict each SELECT statement. EditSQL~\cite{zhang2019editing}, IGSQL~\cite{cai2020igsql} and R$^2$SQL~\cite{hui2021dynamic} consider the dialogue context during translating the utterance into SQL query. GAZP~\cite{zhong-etal-2020-grounded} proposes a zero-shot method to adapt an existing semantic parser to new domains. PIIA~\cite{li2020you} proposes a human-in-loop method to enhance Text-to-SQL performance.

\noindent \textbf{Graph Neural Network}
Graph neural network (GNN)~\cite{li2015gated} has been widely applied in various NLP tasks, such as text classification~\cite{chen2020neural,lyu2021let}, text generation~\cite{zhao2020line}, dialogue state tracking~\cite{chen2020schema,zhu2020efficient} and dialogue policy~\cite{chen2018policy,chen2018structured,chen2019agentgraph,chen2020distributed,chen2020structured}.
It also has been used to encode the schema in a more structured way. Prior work ~\cite{bogin2019representing} constructed a directed graph of foreign key relations in the schema and then got the corresponding schema representation with GNN. Global-GNN ~\cite{bogin2019representing} also employed a GNN to derive the representation of the schema and softly select a set of schema nodes that are likely to appear in the output query. Then, it discriminatively re-ranks the top-K queries output from a generative decoder. We proposed Graph Projection Neural Network (GPNN), which is able to extract the abstract representation of the NL question and the semantic schema.

\noindent \textbf{Generalization Capability}
To improve the compositional generalization of a sequence-to-sequence model, SCAN~\cite{lake2018generalization} (\textbf{S}implified version of the \textbf{C}omm\textbf{A}I \textbf{N}avigation tasks) dataset has been published. SCAN task requires models to generalize knowledge gained about the other primitive verbs (``walk", ``run" and ``look") to the unseen verb ``jump". \citet{russin2019compositional} separates syntax from semantics in the question representation, where the attention weight is calculated based on syntax vectors but the hidden representation of the decoder is the weight sum of the semantic vectors. Different from this work, we look at the semi-structured schema from two perspectives (schema structure and schema semantics). Our proposed GPNN aims to use the schema semantics as the bridge to get abstract representation of the question and schema.

\section{Conclusion}
In this paper, we propose a graph project neural network (GPNN) to abstract the representation of question and schema with simple attention way. We further unify the abstract representation of question and schema outputted from GPNN with relative-aware transformer (RAT). The experiments demonstrate that our proposed ShadowGNN can get excellent performance on the challenging Text-to-SQL task. 
Especially when the annotated training data is limited, our proposed ShadowGNN gets more performance gain on exact match accuracy and convergence speed. The ablation studies further indicate the effectiveness of our proposed GPNN. Recently, we notice that some Text2SQL-specific pretrained models have been proposed, e.g., TaBERT~\cite{yin2020tabert} and GraPPa~\cite{yu2020grappa}. In future work, we will evaluate our proposed ShadowGNN with these adaptive pretrained models. 

\section*{Acknowledgements}
We thank the anonymous reviewers for their thoughtful comments.  This work has been supported by No. SKLMCPTS2020003 Project.

\bibliography{anthology,custom}
\bibliographystyle{acl_natbib}

\clearpage
\appendix



\end{document}